\pdfoutput=1

\documentclass[11pt]{article}

\usepackage{acl}
\usepackage{times}
\usepackage{latexsym}
\usepackage{amsmath,graphicx,csquotes,mathtools,hyperref}
\usepackage{multirow,booktabs,colortbl,subcaption}
\usepackage{soul}
\usepackage{xcolor}

\usepackage[T1]{fontenc}

\usepackage[utf8]{inputenc}

\usepackage{microtype}

\newcommand{\sara}[1]{{\textcolor{black}{#1}}}
\newcommand{\alin}[1]{{\textcolor{black}{#1}}}

\title{Dodging the Data Bottleneck: \\ Automatic Subtitling  with Automatically Segmented ST Corpora}

\author{Sara Papi\textsuperscript{$\clubsuit$,$\diamondsuit$}, Alina Karakanta\textsuperscript{$\clubsuit$,$\diamondsuit$}, Matteo Negri\textsuperscript{$\clubsuit$}, Marco Turchi\textsuperscript{$\spadesuit$*} \\
  \textsuperscript{$\clubsuit$}Fondazione Bruno Kessler \\
  \textsuperscript{$\diamondsuit$}University of Trento \\
  \textsuperscript{$\spadesuit$}Zoom Video Communications \\
  \texttt{\{spapi, akarakanta, negri\}@fbk.eu, marco.turchi@zoom.us}
  }

\begin{document}
\maketitle
\begin{abstract}
\newcommand\blfootnote[1]{%
  \begingroup
  \renewcommand\thefootnote{}\footnote{#1}%
  \addtocounter{footnote}{-1}%
  \endgroup
}
\blfootnote{* Work done when working at FBK.}
Speech translation for subtitling
(SubST) is the task of automatically translating speech data into well-formed 
subtitles by inserting subtitle breaks compliant to specific displaying guidelines. 
Similar to speech translation (ST), model training requires parallel data comprising audio inputs paired with their textual 
translations. 
In SubST, however, the text has to be also annotated with subtitle breaks.
So far, this requirement has represented a bottleneck for system development, as confirmed by the dearth of publicly available SubST corpora.
To fill this gap, we propose a method to convert existing ST corpora into SubST resources without human intervention. We build a segmenter model that automatically segments 
texts
into proper subtitles by exploiting audio and text in a multimodal fashion, achieving high segmentation quality in zero-shot conditions.
Comparative experiments with SubST systems respectively trained on manual and automatic segmentations result in similar performance, showing the effectiveness of our approach.
\end{abstract}

\section{Introduction}
\label{sec:intro}


Massive amounts of audiovisual content are available online, and this abundance is accelerating with the spread of online communication during the COVID-19 pandemic. The increased production of pre-recorded lectures, presentations, tutorials and other audiovisual products raises an unprecedented demand for subtitles in order to facilitate comprehension and inclusion of people without access to the source language speech. 
To keep up with such a demand, automatic solutions are seen as a useful support to the limited human workforce of trained professional subtitlers available worldwide \cite{Tardel_2020}.
%
%
Attempts to automatise subtitling have focused on Machine Translation for translating human- or automatically-generated source language subtitles  \cite{
Volk-et-al-2010,Etchegoyhen-2014-sumat,matusov-etal-2019-customizing,koponen-2020}. 
%
%
%
Recently, direct ST systems \cite{berard_2016,weiss2017sequence} have been shown to achieve high performance while generating the translation in the target language without intermediate transcription steps. 
For automatic subtitling, \citet{karakanta-etal-2020-42} suggested that, by directly generating target language subtitles from the audio (i.e. predicting  subtitle breaks together with the translation), the model can improve subtitle segmentation by exploiting additional information like pauses and prosody.
However, the scarcity of SubST corpora makes it hard to build competitive systems for automatic subtitling, especially if no corpus is available for specific languages/domains.

One solution to the SubST data bottleneck could be leveraging ST corpora by inserting subtitle breaks on their target side. Automatic segmentation of a text into subtitles is normally implemented with rule-based approaches and heuristics,
e.g. a break is inserted before a certain length limit is reached.
More involved algorithms (SVM, CRF, seq2seq) predict breaks using a segmenter model trained on subtitling data for a particular language \cite{alvarez-et-al-2016,Alvarez17improving,karakanta-etal-2020-point}. Still, the performance of these models relies on high-quality segmentation annotations for each language, which web-crawled subtitling corpora like OpenSubtitles \cite{lison-etal-2018-opensubtitles2018} rarely contain.

In this work, we address the scarcity of SubST corpora by developing a multimodal segmenter\footnote{Code and model available at \url{https://github.com/hlt-mt/FBK-fairseq}.} able
to automatically annotate existing ST corpora with subtitle breaks in a zero-shot fashion.
Specifically, our segmenter exploits, for the first time in this scenario, the source language audio (here: en) and segmented target text already available in a few languages 
(here: de, en, fr, it).
Its key strength is the ability to segment not only target languages for which high-quality segmented data is available but also unseen languages 
having some degree of similarity with those covered by the original ST resource(s).
This opens up the possibility to automatically 
obtain synthetic SubST training data for previously not available languages.
%
%
%
Along this direction, our zero-shot segmentation results on two unseen languages 
(es, nl)
show that training a SubST system on automatically-segmented data leads to comparable performance compared to using a gold, manually-segmented corpus.

 \section{Methodology}
\label{sec:method}
Our method to leverage ST corpora for SubST can be summarized as follows:
\emph{i)} we train different segmenters
on available human-segmented subtitling data to select the best performing one;
\emph{ii)} we run the selected segmenter in a \emph{zero-shot} fashion (i.e. without fine-tuning or adaptation) to insert subtitle breaks in unsegmented text data of \textit{unseen} languages;
\emph{iii)} then, the automatically annotated texts are paired with the corresponding audio to obtain a synthetic parallel SubST corpus; 
\emph{iv)} finally, a SubST model is trained on the synthetic corpus.

We test our method on two language 
pairs (en-es, en-nl)
by comparing the results of  SubST models trained on  synthetic data with those of 
identical models trained on
original gold data.

\subsection{Segmenter}


We adopt the general segmentation approach of \cite{karakanta-etal-2020-must} where a sequence to sequence \textit{Textual segmenter}, trained on pairs of unsegmented-segmented text, takes unsegmented text as input and inserts subtitle breaks.

\begin{figure}[htb]
    \centering
    \includegraphics[width=0.48\textwidth]{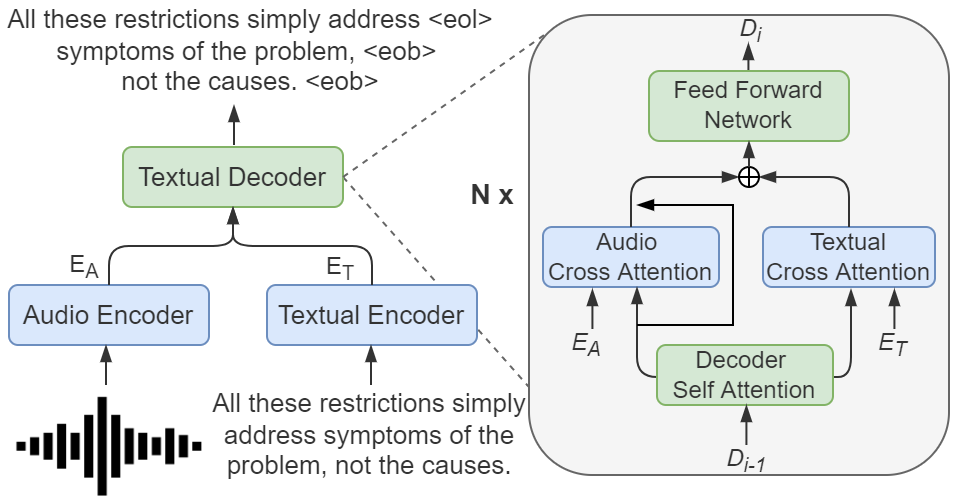}
    \caption{Parallel Multimodal segmenter architecture.}
    \label{fig:multimodal}
\end{figure}

To improve segmentation quality, we extend this approach in two ways. Our first extension is multimodal training. Since speech phenomena, such as pauses and silences, can strongly influence the structure of the subtitles~\cite{Carroll-Ivarsson-98}, we expect that 
information from the speech modality 
could improve segmentation.
To explore this hypothesis, we extend the textual segmenter with a multimodal architecture \cite{sulubacak2020multimodal}, which receives input from different modalities: in our case, audio and text.\footnote{Images and videos with subtitling material are often protected by copyright and thus not publicly available. Improving the segmenter with data from the visual modality is thus left to future work depending on the availability of such resources.} 
Our \textit{Multimodal segmenter} is built using an architecture with two encoders: one for the text (with the same structure as the textual 
segmenter) and one for the audio.
We combine the encoder states obtained by the two encoders using parallel cross-attention \cite{bawden-etal-2018-evaluating},\footnote{We also tested sequential cross-attention \cite{zhang-etal-2018-improving} but do not report results since they are slightly worse compared to parallel cross-attention.} as it proved to be effective both in speech and machine translation \cite{kim-etal-2019-document,Gaido2020}. 
Parallel attention (Figure \ref{fig:multimodal}) is computed by attending at the same intermediate representation (the decoder self-attention); then, the audio encoder cross-attention and the text encoder cross-attention are summed together and fed to the feed-forward layer.

Since subtitling constraints are the same across several languages, our second extension is to learn segmentation multilingually.
To this aim, we follow standard approaches used in MT and ST, respectively (see Appendix \ref{sec:exp} for more details): for the textual segmenter we combine 
samples from multiple languages in the same training 
step \cite{ott2018scaling}; for
the multimodal segmenter
we add a prefix language token to the target text \cite{inaguma2019multilingual}. 
As in MT \cite{ha2toward}, multilingual training in ST has been shown to enhance  
performance \cite{wang-etal-2020-fairseq} while 
maintaining only one model for multiple languages.

\subsection{Data, Baselines and Evaluation}
\label{subsec:data_eval}
\textbf{Data.} 
To train
our textual and multimodal segmenters,
we use en$\rightarrow$\{de/fr/it\} sections of
MuST-Cinema \cite{karakanta-etal-2020-must}, the only publicly available SubST dataset. More details about the data selection are provided in Appendix \ref{sec:exp}.
To 
test the segmenters in zero-shot conditions
(Section \ref{sec:unseen})
and train our SubST models (Section \ref{sec:subst-synt-data}),
we select two target languages also contained in 
MuST-Cinema:\footnote{Though
present in MuST-Cinema, es and nl data are 
only used for testing purposes so as to simulate the zero-shot conditions required 
to select the best segmenter and evaluate our SubST systems.
}
Dutch (an SOV -- Subject-Verb-Object -- language) and Spanish 
(SVO).
Using the corpus notation, subtitle breaks are defined as: \textit{block break} \texttt{<eob>}, which marks the end of the current subtitle displayed on screen, and \textit{line break} \texttt{<eol>}, which splits consecutive lines inside the same block.

\noindent
\textbf{Baselines.} We compare the performance of the segmenters with two baselines. One is a rule-based method (\textit{Count Chars}) where a break is inserted before a 42-character limit. 
This is the simplest method to always produce length-conforming subtitles and serves as a lower bound for segmentation performance. Our second baseline (\textit{Supervised}) is a neural textual segmenter trained on OpenSubtitles, the largest collection of publicly available textual subtitling data, for the respective language (es, nl).
Although OpenSubtitles is available for a variety of languages, it has some limitations: it does not contain audio, the subtitle and segmentation quality varies since subtitles are often machine-translated or created by non-professionals, and line breaks were lost when pre-processing the subtitles to create the corpus. These limitations may have a detrimental effect on the quality of segmenters trained on this data \cite{Karakanta2019AreSC}. Complete details on experimental settings are presented in Appendix \ref{sec:exp}.

\noindent 
\textbf{Evaluation.} 
To evaluate both the quality of the SubST output and the accuracy of our segmenters, we resort to reference-based evaluation. For translation quality of the SubST output 
we use BLEU \cite{post-2018-call}\footnote{BLEU+c.mixed+\#.1+s.exp+tok.13a+v.1.5.1}, computed on the text from which the subtitle breaks are removed.
For segmentation accuracy, we use $Sigma$~\cite{karakanta-etal-2022-segmentation}, a novel subtitle segmentation metric based on BLEU. Sigma is the ratio of the segmentation achieved for a given text to the best segmentation that could be achieved. Contrary to other standard segmentation metrics, such as F1, it can be computed when the output text is different than the reference text. 
%
To ensure that the system does not over- or under-generate subtitle breaks, we additionally report \emph{Break coverage} 
computed as follows:
\setlength{\jot}{10pt}
\begin{equation*}
    \footnotesize
    Coverage(\%) = \left( \frac{\# \texttt{<break>}_{pred}}{\# \texttt{<break>}_{ref}} \cdot 100 \right) - 100 
\end{equation*}
where \texttt{<break>} corresponds to either  \texttt{<eol>} or \texttt{<eob>}. 
EOL and EOB coverage obtains negative values when the segmenter inserts less breaks than required or positive values when it inserts more.
Lastly, we use
\emph{length conformity} (or characters per line -- CPL), 
corresponding to the percentage of subtitles not exceeding the allowed maximum length of 42 
CPL, as per TED guidelines.\footnote{\url{https://www.ted.com/participate/translate/subtitling-tips}}

\begin{table*}[!ht]
\centering
\small
\setlength{\tabcolsep}{5pt}
\begin{tabular}[width=\textwidth]{lc|cc|cc|cc|cc} 
\bottomrule
\multirow{2}{*}{Segmenter} & \multirow{2}{*}{Training} & \multicolumn{2}{c|}{English} & \multicolumn{2}{c|}{French} & \multicolumn{2}{c|}{German} & \multicolumn{2}{c}{Italian}  \\ 
\cline{3-10}
& \multicolumn{1}{c|}{}                          & Sigma & CPL                   & Sigma & CPL                  & Sigma & CPL                  & Sigma & CPL                   \\ 
\hline
Count Chars & & 63.71 & 100\% & 62.87 &100\% & 62.34 & 100\% & 61.49 & 100\% \\ \hline
\multirow{2}{*}{Textual} & mono & 84.87 & \textbf{96.6\%} & 83.68 & \textbf{96.7\%} & 83.62 & 90.9\% & 82.22 & 90.0\% \\
& multi & 85.98 & 88.5\% & 84.56 & 94.3\% & 84.02 & 90.9\% & 83.04 & \textbf{91.2\%} \\
\hline
\multirow{2}{*}{Multimodal} & mono & 85.76 & 94.8\% & 84.25 & 93.9\% & 84.22 & \textbf{91.4\%} & 82.62 & 89.9\% 
\\
    & multi & \textbf{87.44} & 95.0\% & \textbf{86.49} & 94.1\% & \textbf{86.4} & 89.9\% & \textbf{85.33} & 90.0\% \\
\specialrule{.1em}{.05em}{.05em} 
\end{tabular}
    \caption{Segmentation results on \textit{seen} languages.}
    \label{tab:segmenter_results}
\end{table*}

\section{Segmentation on \emph{seen} languages}
\label{sec:seen}

We train the mono/multi-lingual versions of our \textit{Textual/Multimodal} segmenters for the four languages (de, en, fr, it) and measure their performance in terms of Sigma and CPL. The results are shown in Table \ref{tab:segmenter_results}. 

Looking at the Sigma values, both the \textit{Textual} and the \textit{Multimodal} segmenter perform better than the rule-based baseline, despite a small drop in CPL.
The \textit{Multimodal} segmenter always outperforms the \textit{Textual} one by 2 Sigma points on average 
and inserts break symbols more accurately. 
Moreover, it benefits from multilingual training on all languages.
In contrast, overall subtitle conformity is higher for the \textit{Textual} segmenter in 3 out of 4 languages, where its CPL scores are 1.2-2.6 percentage points above those obtained by the \textit{Multimodal} one. In addition, except for one case (German), higher CPL values are obtained 
with monolingual training
.

\section{Zero-shot segmentation }
\label{sec:unseen}
Aiming to build a SubST model for unseen languages (es, and nl), we first select the best segmenter for generating synthetic en$\rightarrow$es/nl data.
%
%
%
As shown in Table \ref{tab:zero_shot}, all the models that receive only text as input (\textit{Count Chars}, \textit{Supervised} and \textit{Textual}) achieve low segmentation performance, with Sigma ranging between 63-75. The zero-shot \textit{Textual} segmenter achieves higher segmentation quality compared to the \textit{Count Chars} and \textit{Supervised} baselines by 10 points. However, its main drawback is the inability to copy the actual text, as shown by the BLEU values of 61 for nl and 69 for es. In this respect, the baselines perform much better. Despite being trained on subtitling data for the particular language, the low segmentation performance of \textit{Supervised} can be attributed to the different domain compared to the MuST-Cinema test set. For example, MuST-Cinema mainly contains long sentences with multiple breaks, while in OpenSubtitles we rarely come across sentences with more than three breaks. 
Moreover, both \textit{Supervised} and \textit{Textual} generate subtitles conforming to the CPL constraint in only 
70\% of the cases, despite 
having received only length-conforming subtitles as training data. 
The negative values of EOL and EOB coverage show that all textual methods under-generate subtitle breaks. 
From these results we can conclude that zero-shot segmentation 
does not perform satisfactorily with textual input only.

\begin{table}[htb]
\centering
\small
\setlength{\tabcolsep}{3pt}
\begin{tabular}{l|ccccc} 
\specialrule{.1em}{.05em}{.05em} 
\multicolumn{6}{c}{Dutch} \\ 
\hline
Segmenter & BLEU & Sigma & CPL & EOL & EOB \\ 
\hline
\multicolumn{1}{l|}{Count Chars}& \textbf{100} & 63.2 & \textbf{100\%} & \textbf{-21.2\%} & -7.1\% \\
\hline
\multicolumn{1}{l|}{Supervised}& 89.5 & 64.4 & 71.2\% & -31.4\% & -51.3\% \\
\hline

\multicolumn{1}{l|}{Textual}& 61.3 & 74.4 & 77.8\% & -23.4\% & -9.9\% \\
\multicolumn{1}{l|}{Multimodal}& 99.9 & \textbf{80.3} & 91.4\% & -27.2\% & \textbf{+0.4\%} \\
\specialrule{.1em}{.05em}{.05em} 
\multicolumn{6}{c}{Spanish} \\ 
\hline
Segmenter& BLEU & Sigma & CPL & EOL & EOB \\ 
\hline
\multicolumn{1}{l|}{Count Chars}& \textbf{100} & 63.2  & \textbf{100\%} & -24.6\% & \textbf{-4.4\%} \\
\hline
\multicolumn{1}{l|}{Supervised} & 92.6 & 64.1  & 71.2\% & -32.3\% & -45.4\% \\
\hline
\multicolumn{1}{l|}{Textual}& 69.6 & 75.8 & 70.1\% & -47.6\% & -19.3\% \\
\multicolumn{1}{l|}{Multimodal}& 99.6 & \textbf{78.7} & 91.8\% & \textbf{-22.4\%} & +4.7\% \\\bottomrule
\end{tabular}
\caption{Segmentation results on \textit{unseen} languages.}
\label{tab:zero_shot}
\end{table}


In comparison, the \textit{Multimodal} segmenter performs significantly better.
It reaches an absolute gain of 6.1 Sigma points 
for nl and 2.9 
for es compared to \textit{Textual}. Moreover, contrary to \textit{Textual} and \textit{Supervised}, the \textit{Multimodal} model learnt to perfectly copy the text, as shown by the high
BLEU scores (up to 99.9 on nl), close to the maximum score of a method -- \textit{Count Chars} -- that by design does not change the original text. 
The CPL results are in agreement with BLEU: for both languages, the \textit{Multimodal} model respects the length constraint in more than 91\% of the subtitles. 
Strikingly, even if the two target languages were never seen by the model, these results are similar to those obtained on seen languages (see Table \ref{tab:segmenter_results}).
Unlike the rest of the models, \textit{Multimodal} is the only model which does not under-generate \texttt{<eob>}. 
This is in line with the results of
\citet{karakanta-etal-2020-42}, who showed that exploiting the audio in ST is beneficial for inserting subtitle breaks (\texttt{<eob>}, for instance, typically corresponds to longer speech pauses). 
The results are more discordant for the EOL Coverage. On es, \textit{Multimodal} shows a lower tendency to under-generate, while on nl both models fail to insert at least the 23.4\% of \texttt{<eol>}.
We assume this phenomenon is caused by the lower frequency of \texttt{<eol>} in the corpus, since a subtitle can be composed of only one line, as well as by the higher difficulty in placing the break for which the system cannot resort to speech clues (e.g. pauses).
%
%
%
%
%

\paragraph{Ablation.} To test the effectiveness of the \textit{Multimodal} model also in the absence of similar languages in the training set, we train it on a limited set of Latin languages (Italian and French) and test it on Dutch, which is a Germanic language.

The results (\textit{fr, it only}) are shown in Table \ref{tab:ablation}. 
Even if trained on 
only two languages 
from a different language group
, the \textit{fr, it only Multimodal} model shows competitive results. In terms of segmentation, there is only a slight degradation of 3 Sigma points compared to the full multilingual \textit{Multimodal} model and a 3.6\% drop in CPL conformity, which could be attributed to a lower EOL coverage. However, it is still significantly better in terms of Sigma, CPL conformity and EOB coverage compared to all the other segmenters (\emph{Count Chars}, \emph{Supervised}, and \emph{Textual}). 
In terms of changes to the text, as show by BLEU, it is on par with 
\textit{Supervised}, a model trained only on Dutch subtitles, and better than the \textit{Textual} by 25 BLEU points. The presence of related languages seems to help the model better copy the text, since the main drop compared to the full \textit{Multimodal} model is in terms of BLEU.  
Overall, we can conclude that the presence of related languages in the training set can enhance the performance, but the segmentation accuracy and conformity are only minimally affected. The results obtained by the \textit{fr, it only Multimodal} confirm the ability and superiority of this model in segmenting texts on unseen languages also belonging to different language groups.

\begin{table}[htb]
\centering
\small
\setlength{\tabcolsep}{3pt}
\begin{tabular}{l|ccccc} 
\bottomrule
Segmenter & BLEU & Sigma & CPL & EOL & EOB \\ 
\hline
\multicolumn{1}{l|}{Count Chars}& \textbf{100} & 63.2 & \textbf{100\%} & \textbf{-21.2\%} & -7.1\% \\
\multicolumn{1}{l|}{Supervised}& 89.5 & 64.4 & 71.2\% & -31.4\% & -51.3\% \\
\multicolumn{1}{l|}{Textual}& 61.3 & 74.4 & 77.8\% & -23.4\% & -9.9\% \\
\hline
\multicolumn{1}{l|}{Multimodal}& 99.9 & \textbf{80.3} & 91.4\% & -27.2\% & \textbf{+0.4\%} \\
\multicolumn{1}{l|}{ \quad - \textit{fr, it only}}& \textit{88.9} & \textit{77.0} & \textit{87.8\%} & \textit{-34.8\%} & \textit{\textbf{-0.4\%}} \\
\toprule
\end{tabular}
\caption{Ablation results on MuST-Cinema amara en$\rightarrow$nl. All but the last line are from Table \ref{tab:zero_shot}.}
\label{tab:ablation}
\end{table}

\paragraph{Limitations.}
So far, our results
indicate
the effectiveness of \textit{Multimodal} segmentation to automatically 
turn existing ST corpora into 
SubST 
training data.
In addition,
at least for the Western European languages considered in our experiments, our approach can be successfully applied in zero-shot settings
involving 
languages not present in the training data
which also belong to different language groups. 
Not being possible to verify due to the lack of suitable benchmarks,
the possibility of porting our approach to scenarios involving different alphabets is not verified in this work.
This would require, at least, a vocabulary adaptation which represents a well-known problem in multilingual
approaches to MT/ST 
\cite{garcia-etal-2021-towards}.
%
Nevertheless, even in the worst case in which some degree of similarity across languages is required for zero-shot automatic segmentation, 
we believe that these results  indicate a viable path towards overcoming the scarcity of SubST resources. In the next section, we will test this hypothesis.


\section{SubST with Synthetic Data}
\label{sec:subst-synt-data}
%
%
%
%
%
%
%
%
%
%
%
%
%
Since 
our \textit{Multimodal}
segmenter 
achieves the best performance overall, we use 
it to automatically 
generate
the synthetic counterpart of the 
en$\rightarrow$\{es, nl\} sections of 
MuST-Cinema. 
The resulting data is respectively used to train two SubST  
systems.
The goal is to achieve comparable performance to that of similar
models trained
on manually segmented subtitles. 
For this purpose, using the same architecture, we 
also 
train two 
systems
on the original manual segmentations of MuST-Cinema.
%
%
%
%

\begin{table}[htb]
\centering
\small
\setlength{\tabcolsep}{4pt}
\begin{tabular}{l|ccccc} 
\specialrule{.1em}{.05em}{.05em} 
\multicolumn{6}{c}{Dutch} \\ 
\hline
Data & BLEU & Sigma & CPL & EOL & EOB \\ 
\hline
\multicolumn{1}{l|}{Original} & \textbf{25.3}* & \textbf{81.58} & 91.2\% & -36.8\% & +8.0\% \\
\multicolumn{1}{l|}{Synthetic} & 24.3* & 75.52 & \textbf{94.7\%} & \textbf{-20.4\%} & \textbf{+4.8\%} \\
\specialrule{.1em}{.05em}{.05em} 
\multicolumn{6}{c}{Spanish} \\ 
\hline
Data & BLEU & Sigma & CPL & EOL & EOB \\ 
\hline
\multicolumn{1}{l|}{Original} & \textbf{30.7}* & \textbf{79.21} & \textbf{96.7\%} & \textbf{-10.0\%} & +10.9\% \\
\multicolumn{1}{l|}{Synthetic} & \textbf{30.7}* & 77.84 & 94.2\% & -21.5\% & \textbf{+9.9\%} \\ \bottomrule
\end{tabular}
\caption{Results of the SubST systems. 
The * stands for statistically \textbf{not} significant results according to the bootstrap resampling test \cite{koehn-2004-statistical}.
}
\label{tab:final_res}
\end{table}

As shown in Table \ref{tab:final_res}, the SubST system trained on our automatically segmented data (\textit{Synthetic}) shows comparable performance with the system trained on the original segmentation (\textit{Original}).
The BLEU between the two models is identical for es, while for nl the difference is not significant.
On the contrary, the Sigma for the system trained on manual segmentations is higher than for the synthetic ones by 6 points for nl but less than 2 for es.
These results highlight that the breaks introduced by a non-perfect automatic segmentation influence the way the subtitle breaks are placed in the translation but not necessarily the translation itself.
For the length constraint, both systems obtain high CPL conformity, with the \textit{Synthetic} model scoring 3.5\% more on nl and 2.5\% less on es. 
This is related to the number of \texttt{<eol>} and \texttt{<eob>} 
inserted by the system:
the more subtitle breaks are present, the more fine-grained is the segmentation, leading to higher conformity. 
Indeed, CPL is 
higher when the Break Coverage is high. 

\paragraph{Manual Analysis.}
Upon 
examination of the segmentation patterns of the two 
en$\rightarrow$es systems,\footnote{We were unable to replicate the analysis on nl as we do not have the required linguistic competences.}
we did not identify particular 
differences.
Specifically, the inserted \texttt{<eob>} tags follow punctuation marks in 76\% of the cases for both models and are followed by prepositions and conjunctions in 32\% and 29\% for \textit{Original} and \textit{Synthetic} respectively.
Similar patterns between outputs were observed for \texttt{<eol>} too, which is followed by a comma in the majority of cases and by the same function words as \texttt{<eob>}. 
These results suggest 
that systems trained on automatically segmented data are able to reproduce similar segmentation patterns to those trained on original data without showing a significant degradation in the translation.

\section{Conclusions}
We presented 
an automatic 
segmenter able to turn existing ST corpora into SubST 
training data. Through comparative experiments on two language pairs in zero-shot conditions, we showed that SubST systems trained on this synthetic data are competitive with those built on human-annotated subtitling corpora. Building on these positive results, and conditioned to the availability of suitable benchmarks, verifying the portability of the approach to a larger set of languages and domains is our priority for future work.


\bibliography{anthology,custom}
\bibliographystyle{acl_natbib}

\appendix


\section{Experimental Settings}
\label{sec:exp}

\subsection{Data Selection}
For the initial experiments aimed to train textual and multimodal segmenters and to select the best one (step 1 of the process described in Section \ref{sec:method}), we use three sections of MuST-Cinema \cite{karakanta-etal-2020-must}, the only \alin{to date} publicly available SubST dataset,\footnote{\url{https://ict.fbk.eu/must-cinema/} - License: CC BY-NC-ND 4.0} namely \sara{French, German, and Italian.}
Each section contains paired audio utterances, \sara{English} transcripts, and translations \sara{in the corresponding language}, where both sides of the text are built from subtitles \sara{created by humans.}
For French (275K sentences), German (229K sentences) and Italian (253K sentences), we 
\sara{collect the segmented translations of}
the corresponding MuST-Cinema sections. 
For English,
\sara{we concatenate the segmented transcripts of the previous three sections (757K sentences).}
\sara{For each language (de, en, fr, it), the training data for the segmenter consists of unsegmented texts and, in the case of the multimodal segmenter, also audio as the source input, and of segmented texts (subtitles) as the target.}
Using the corpus notation, subtitle breaks are defined as: \textit{block break} \texttt{<eob>}, which marks the end of the current subtitle displayed on screen, and \textit{line break} \texttt{<eol>}, which splits consecutive lines inside the same block. 
\sara{For unsegmented texts, \texttt{<eob>} and \texttt{<eol>} are removed.}

\subsection{Systems}
We use the Adam optimizer and inverse square-root learning rate (lr) scheduler for all the trainings.

The \textit{Textual} segmenter is a Transformer-based \cite{transformer} architecture consisting of
3 encoder layers and 3 decoder layers.
We set the hyper-parameters as in the fairseq \cite{ott-etal-2019-fairseq} multilingual translation task, both for the mono- and multilingual textual segmenters. 
For the multilingual model, a mini-batch for each language direction (here: 4) is built and the model weights are updated after each mini-batch, a mechanism already present in fairseq Multilingual Machine Translation \cite{ott-etal-2019-fairseq}.

The \textit{Multimodal} segmenter is an extension of the textual segmenter encoder-decoder structure with an additional speech encoder composed of 12 Transformer encoder layers as in the original speech-to-text task \cite{wang-etal-2020-fairseq}
but with the addition of a CTC \cite{10.1145/1143844.1143891} module to avoid the speech encoder pre-training \cite{gaido-etal-2021-ctc}. The training of the multilingual models is performed by pre-pending the language token (\texttt{en}, \texttt{de}, \texttt{fr}, \texttt{it}) to the target sentence 
\cite{inaguma2019multilingual}, a mechanism 
already present in the Fairseq Speech-to-text library \cite{wang-etal-2020-fairseq}.
The encoder and decoder embeddings are shared.
We select the hyper-parameters of the original implementation,\footnote{\url{https://github.com/pytorch/fairseq/blob/main/examples/speech_to_text/docs/mustc_example.md}} 
except for a higher learning rate 
of $1\cdot10^{-3}$, 
since 
 pre-training was skipped. 
The vocabulary is generated using SentencePiece \cite{kudo-richardson-2018-sentencepiece}, setting the size to 10k unigrams both for the mono- and multilingual segmenters. 

For the \textit{Supervised} baseline using OpenSubtitles data, we follow the data selection process for the highest-performing segmenter in \cite{karakanta-etal-2020-point} (\textit{OpenSubs-42}). We first filter sentences with subtitles of maximum 42 characters. Since line breaks are not present in OpenSubtitles, we substitute \texttt{<eob>} symbols with \texttt{<eol>} with a probability of 0.25, paying attention not to insert two consecutive \texttt{<eol>}. 
This proportion reflects the \texttt{<eol>}/\texttt{<eob>} distribution featured by  the MuST-Cinema training set.
We noted that almost 90\% of the sentences filtered contain only one subtitle. This is not very informative for the segmenter, since the only operation required is inserting one \texttt{<eob>} at the end of the sentence. For this reason, we further select only sentences with at least two subtitles (or two subtitle lines). This results in 2,956,207 sentences for es and 683,382 sentences for nl. We then add the same number of sentences containing only one subtitle. After this process, we obtain 5,912,414 sentences for es and 1,366,764 sentences for nl. The supervised baseline is trained with the same settings as the textual monolingual segmenter. 

For the \textit{Count Chars} baseline, a break is inserted 
before reaching the 42-character limit, as per TED guidelines.
If the 42-character limit is reached in the middle of a word, the break is inserted before this word.
This method will always obtain a 100\% conformity to the length constraint. 
As with the data filtering process, \texttt{<eol>} is inserted with probability of 0.25.

For the SubST models discussed in Section \ref{sec:subst-synt-data}, we use the speech-to-text task \textit{small} architecture of fairseq with the additional CTC module as in \cite{papi-etal-2021-dealing}.

We use 4 GPUs K80 for training all the architectures: it takes around 1 day for the textual-only and around 1 week for the multimodal segmenters and the SubST models. 
All results are obtained by averaging 7 checkpoints (best, three preceding and three succeeding checkpoints).

\end{document}